\documentclass[lettersize,journal]{IEEEtran}
\ifCLASSINFOpdf
\else
\fi
\usepackage{graphicx}

\usepackage{algorithm}
\usepackage[noend]{algpseudocode}
\usepackage{algorithmicx}
\usepackage{hyperref}
\usepackage{booktabs}
\usepackage{amsmath}
\usepackage{amssymb}
\begin{document}
%
\title{Network Resource Optimization for ML-Based UAV Condition Monitoring with Vibration Analysis}
%
%
%

\author{Alexandre~Gemayel, Dimitrios~Michael~Manias, and~Abdallah~Shami
\thanks{A. Gemayel, D.M. Manias, and A. Shami are with the Department of Electrical and Computer Engineering at Western University in London, Ontario, Canada. Emails: \{agemayel, dmanias3, Abdallah.Shami\}@uwo.ca.}
}

%
%

\markboth{Accepted in: IEEE Networking Letters}%
{Gemayel \MakeLowercase{\textit{et al.}}: TBD}
%



\maketitle

\begin{abstract}
  As smart cities begin to materialize, the role of Unmanned Aerial Vehicles (UAVs) and their reliability becomes increasingly important. One aspect of reliability relates to Condition Monitoring (CM), where Machine Learning (ML) models are leveraged to identify abnormal and adverse conditions. Given the resource-constrained nature of next-generation edge networks, the utilization of precious network resources must be minimized. This work explores the optimization of network resources for ML-based UAV CM frameworks. The developed framework uses experimental data and varies the feature extraction aggregation interval to optimize ML model selection. Additionally, by leveraging dimensionality reduction techniques, there is a 99.9\% reduction in network resource consumption.
\end{abstract}


%
\IEEEpeerreviewmaketitle

\section{Introduction}
%
%
%
%
\IEEEPARstart{E}{merging} 
Unmanned Aerial Vehicle (UAV) applications, such as Smart Cities, have highlighted the necessity of real-time Condition Monitoring (CM) through Anomaly Detection (AD) and health analytics to ensure operational safety and integrity \cite{mohammed2014uavs}. Given that variations in vibration patterns can signal structural or component damage, vibration analysis is essential to identify adverse conditions in UAVs. \par
Vibration sensors mounted on UAVs produce a wealth of data, used to detect anomalies (\textit{i.e.,} propeller cracks, faulty motors, misaligned components, worn-out bearings, \textit{etc.}). However, the onboard processing and transmission (to the cloud or base station) of sensor data presents operational challenges. \par
Large-scale, cloud-based, and distributed Machine Learning (ML) systems require effective network traffic management when processing massive amounts of data. Network limitations, including bandwidth, latency, and congestion, must be carefully managed when data is transported between sensor, storage, and processing units to ensure operational efficiency. Excessive network traffic can cause bottlenecks that impair the overall system performance through delays in data transfer resulting in stale model insights. Additionally, the communication link between the drone and the base station may face challenges due to non-line-of-sight communication, low transmission power from the drone, interference, poor weather conditions, and several other factors. Poor network resource management can adversely affect model training times and inference speed, leading to data loss, decreased throughput, and higher overhead. To address these challenges, effective network resource management strategies, including data compression and batching are essential. These strategies help ensure ML system responsiveness and model performance. \par
The work presented in this paper explores the effect of network resource optimization on ML model performance for UAV CM. The CM framework is optimized by adjusting the data aggregation scheme to find the ideal data subset size for feature extraction to minimize the network resources consumed while simultaneously ensuring ML model performance. The results include an extensive analysis investigating various subset sizes and their effect on the system. \par
The remainder of this paper is structured as follows. Section II overviews related work. Section III discusses the CM monitoring framework. Section IV presents and analyzes the results. Finally, Section V concludes the paper.

\section{Related Work}

Vibration analysis has been used extensively in UAV CM and AD. Simsiriwong and Sullivan use sensor data from multiple accelerometers on the wing structure of a UAV to determine the vibrational characteristics and dynamic properties of a composite UAV wing through frequency analysis \cite{simsiriwong2012experimental}. Radkowski and Szulim identify and eliminate adverse vibrations appearing in sensor data when a UAV performs aerial maneuvers using vibrational analysis and rotor modelling \cite{radkowski2014analysis}. Wolfram, \textit{et al.} develop models of the UAV drive trains and use various sensors yielding differential pressure, current, voltage, sound, thermocouple, and acceleration data \cite{wolfram2018condition}. \par
Thresholding methods have also been developed to identify anomalous conditions. Bektash and la Cour-Harbo propose the use of periodograms and power spectrum estimations derived from vibration data to infer the mechanical integrity of the UAV and indicate failures \cite{bektash2020vibration}. Al-Haddad, \textit{et al.} use the Fast Fourier Transform (FFT) to extract frequency-domain features from the sensor data which is then used to identify faults through the isolation and characterization of fault signatures \cite{al2023investigation}. Banerjee, \textit{et al.} perform in-flight detection of vibrational anomalies related to motor defects by performing the FFT on the vibration signal, computing the Power Spectral Density (PSD), binning data into intervals and comparing to a pre-defined threshold value to identify anomalies \cite{banerjee2020flight}. \par
ML methods have also been employed to perform UAV CM. Pourpanah, \textit{et al.} use a fuzzy adaptive resonance neural network and a genetic algorithm for feature selection to extract 15 harmonic features from the vibration signal to identify motor and propellor faults \cite{pourpanah2018anomaly}. Zahra, \textit{et al.} leverage LSTMs for abnormal flight detection and health indicator prediction to estimate the remaining useful life using time-domain signal features, such as RMS, Kurtosis, Skewness, and Crest Factor \cite{zahra2021predictive}. Wang, \textit{et al.}  use LSTMs to perform point AD through uncertainty interval estimation based on 1-dimensional UAV flight sensor data \cite{wang2019data}.

Previous work by the authors considered UAV CM through rotor defect detection in pre/post flight settings \cite{gemayel2024machine}. In past work, a comprehensive analysis was conducted to determine the best-performing ML model and the impact of the various feature sets on the model's predictive performance. However, there was no consideration of the impact of network resource utilization on the optimal solution, and accuracy was the only metric used to compare model performance. To address the limitations of past solutions and the gaps in the state-of-the-art, the work presented in this paper focuses on optimizing both the ML model performance (using additional ML model performance metrics) and the network resource utilization efficiency (using a crafted feature throughput metric).
The contributions of this work are as follows:
\begin{itemize}
    \item The development of a network resource-aware UAV CM framework.
    \item A trade-off analysis between network resource utilization and ML model performance.
\end{itemize}

\section{UAV Condition Monitoring Framework}

\subsection{System Model}

\begin{figure}[!htbp]
\centerline{\includegraphics[width=\columnwidth]{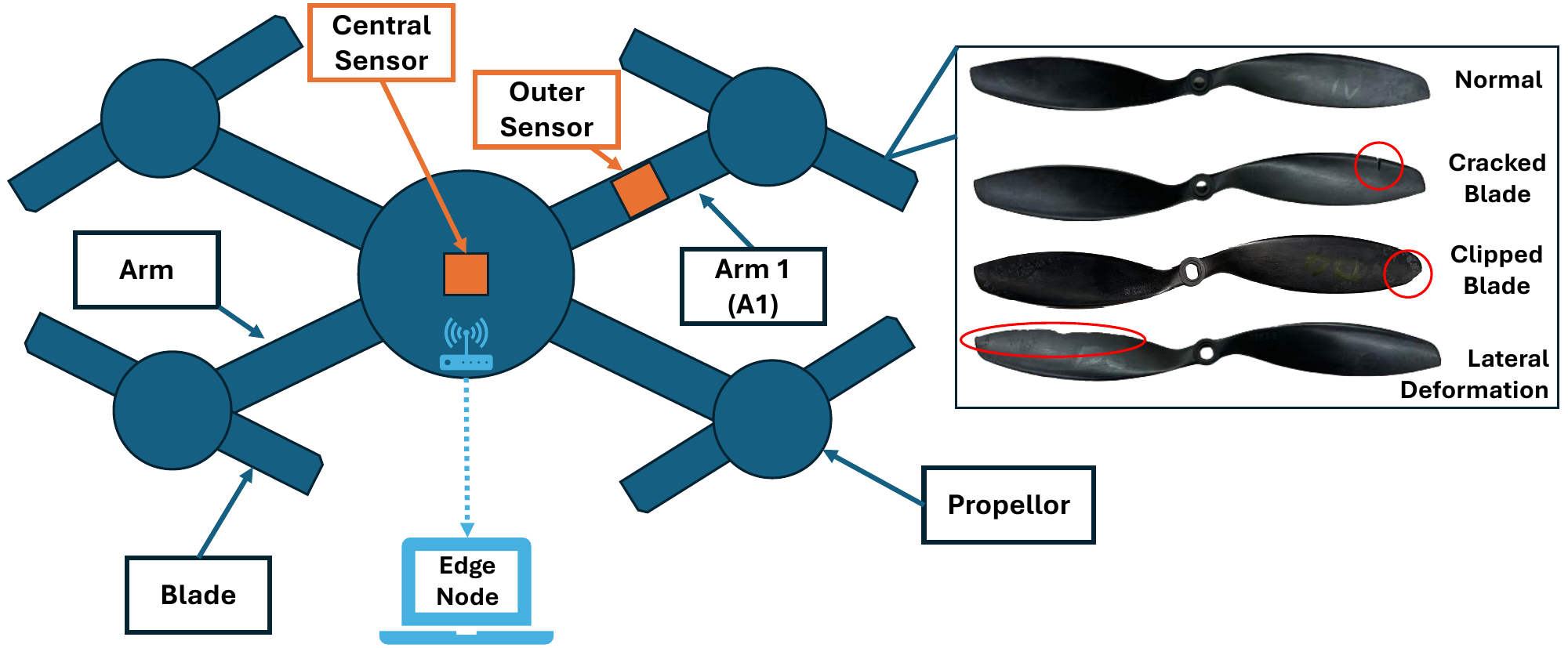}}
\caption{System Model}
\label{system_model}
\end{figure}

Figure \ref{system_model} presents an overview of the system model. In this system, a single drone is used with the intention of expanding to large-scale deployment where the local machine depicted in the diagram will be replaced by a system orchestrator and the single UAV will be scaled up to a network of UAVs. The fully-assembled Helipal Storm 4 drone is used for the proposed experiments and consists of a CC3D, Radio Link of 2.4GHz AT9, Radio System w/ R9D 9-Ch Receiver, and a Li-Po battery with a 14.4V voltage, a current of 2200mAh 35C. The drone is turned on for all of the proposed experiments, and its propellers are active; however, it is stationary and not in flight. This setup represents the pre/post-flight condition. The defect introduced during the experiments is located at the propeller on the UAV's upper-right arm, designated as A1. In this work, seven blades (one normal and six defective) correspond to different types and severities of conditions that might be encountered during flight. The description of these blades is as follows: \textbf{Normal} - No Defect Present, \textbf{Defect 1} - Cracked Blade, Higher Severity, 
\textbf{Defect 2} - Cracked Blade, Lower Severity, \textbf{Defect 3} - Clipped Blade, Higher Severity, \textbf{Defect 4} - Clipped Blade, Lower Severity, \textbf{Defect 5} - Lateral Deformation, Higher Severity, \textbf{Defect 6} Lateral Deformation, Lower Severity. 

Two ADXL345 sensors are used to capture the vibrations. One is positioned in the drone's center, and the other is on its upper-right arm. The ADXL345 is a compact, ultra-low power, 3-axis accelerometer with a 13-bit resolution that can measure up to ±8g. Both the dynamic acceleration caused by motion or shock and the static acceleration of gravity are tracked by ADXL345. Its excellent resolution of 3.9 mg/LSB allows it to measure fluctuations in inclination of less than 1.0°. \par
The SoC microcontroller ESP32, with integrated Wi-Fi 802.11 b/g/n, dual-mode Bluetooth 4.2, and various peripherals, is used. It features two independently controlled cores with separate clock rates up to 240 MHz, and is powered by an Xtensa LX6 CPU produced at 40 nm. There is 520 kB of on-chip SRAM available for executables and data. \par
The ESP32 is connected to a local machine through its built-in Wi-Fi. It gathers UDP packets using Scapy, preventing duplication and confirming checksums for dependable delivery. The script formats the data, groups it, and verifies data integrity while dictating the precise duration for data collection.
\subsection{Experiment Setup}
\begin{figure}[!htbp]
\centerline{\includegraphics[width=\columnwidth]{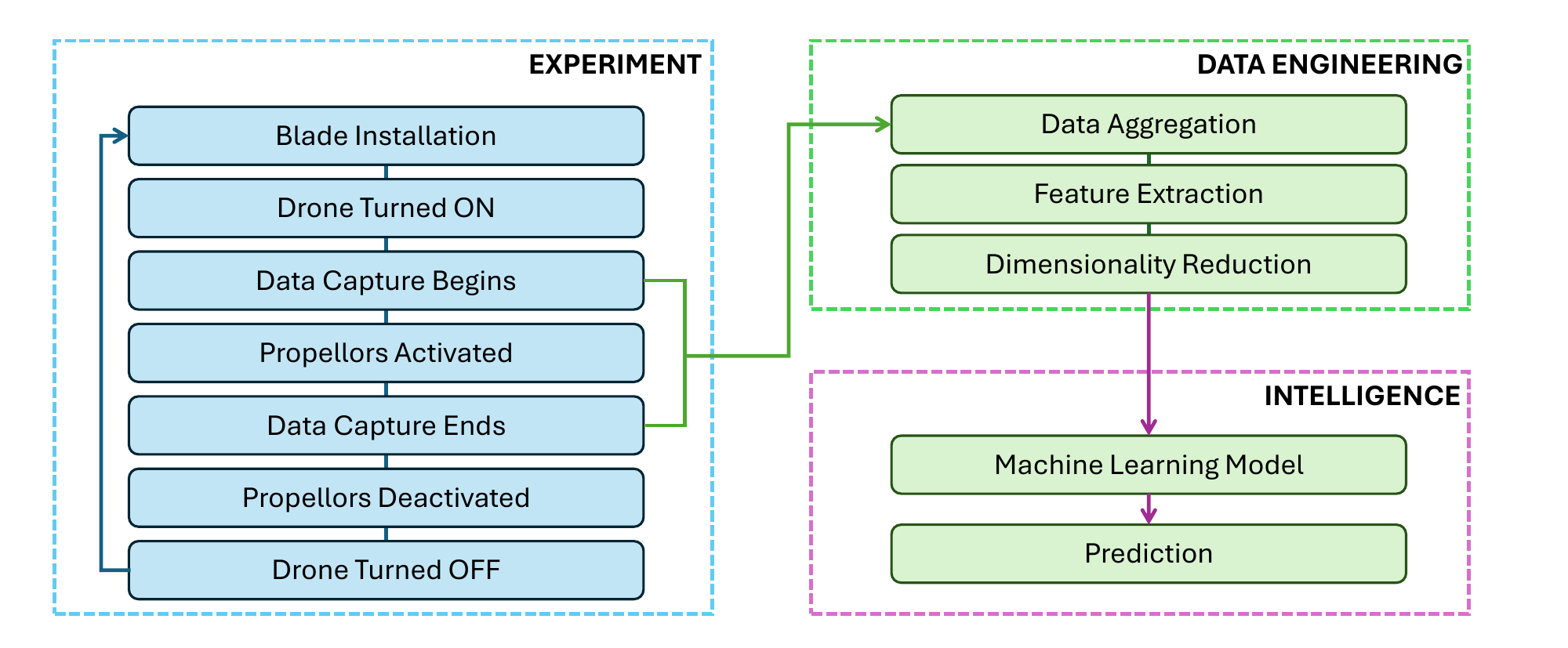}}
\caption{Experiment Overview}
\label{experiment}
\end{figure}
The experiments carried out in this work are similar to the ones conducted in \cite{gemayel2024machine}. A general overview of the experiment methodology is presented in Fig. \ref{experiment}. Each experiment begins by installing the blade on the A1 arm of the drone. It should be noted that all blades were used during the trials. After the blade installation, the drone is turned on. At this point, the data capture begins and the propellors are activated while the drone remains grounded. Data capturing persists for 60 seconds, after which, the propellors are deactivated and the drone is turned off. Eight trials were conducted with the normal non-defective blade. Four trials were conducted for Defect 1, and three trials were conducted for each of the remaining defective blades, resulting in 27 experiments. \par
For each experiment, various aggregation intervals were considered to explore the effect of network resource optimization on ML model performance. Both the central and outer sensors have a frequency of 800 samples/second, meaning that over 60 seconds, 48,000 samples are generated for each sensor, resulting in a total of 96,000 samples. The aggregation intervals (\textit{i.e.}, the number of samples to use for feature extraction) explored along with the associate time interval they represent are listed in Table \ref{agg}.

\begin{table}[]
\caption{Aggregation Intervals and Feature Set Sizes}
\label{agg}
\centering
\begin{tabular}{|c|c|c|}
\hline
\textbf{Number of Samples} & \textbf{Time Interval} & \textbf{Number of Features} \\ \hline
200                        & 250 \textit{ms}                 & 9,320                      \\ \hline
400                        & 500 \textit{ms}                 & 15,476                     \\ \hline
800                        & 1 \textit{s}                    & 27,788                     \\ \hline
1200                       & 1.5 \textit{s}                  & 40,100                     \\ \hline
1600                       & 2 \textit{s}                    & 52,412                      \\ \hline
4000                       & 5 \textit{s}                    & 126,284                     \\ \hline
8000                       & 10 \textit{s}                   & 249,404                     \\ \hline
\end{tabular}
\end{table}

\subsection{Feature Extraction}

Feature extraction occurs on the UAV and the feature set is transmitted to the ML processing unit using the network. The Short Time Fourier Transform (STFT) is calculated for each sensor reading along the X, Y, and Z axes. The wavelet packet transform is used to generate the third feature set. The fourth feature set is calculated using the spectral centroid. The final feature set is the frequency skewness corresponding to the third-order moment. The number of extracted features corresponding to each aggregation interval is listed in Table \ref{agg}. As seen, a longer aggregation interval results in a greater number of features extracted. This is counterintuitive to the objective of this work since it deals with the minimization of the communication resources consumed. To this end, Principle Component Analysis (PCA) is used as a dimensionality reduction technique to reduce excessively large feature sets. 

\subsection{ML Models}
When the features are extracted they are sent to the ML processing unit responsible for determining if the UAV is experiencing an anomaly, specifically a defective propeller blade. The features of each aggregation interval are individual samples and are not dependent on the preceding or following samples. The first stage involves splitting the dataset using a stratified 70-30 train-test split. This means that the distribution of labels across the training and test sets is equal. This step is necessary to ensure that the training environment is consistent with the evaluation/deployment environment, thereby reducing the chances of model drift \cite{manias2023model}. Additionally, dataset splitting is a critical step before additional preprocessing occurs to ensure the test set is completely isolated and unseen for a valid evaluation. After splitting, the data is normalized for all experiments, and PCA is applied to certain feature subsets for some experiments. Since the STFT feature subset is the largest and depends on the aggregation interval, its feature set was selectively targeted for PCA analysis. \par
The ML methods explored as part of this work include the Support Vector Classifier (SVC), the \textit{k}-Nearest Neighbours (KNN), the Decision Tree (DT), and the Random Forest (RF). This subset of models was selected as part of this work as they are lightweight and span a variety of algorithmic methods, including tree-based, ensemble, neighbourhood, and linear. No model fine-tuning was conducted as part of this work. Therefore, all hyperparameters are initialized to default values. 

\subsection{Evaluation Metrics}
This section outlines the metrics used to compare the ML model performance and the network resource consumption. 
\subsubsection{ML Model Performance}
Given the classification nature of the ML problem at hand, standard metrics, including accuracy (ACC), precision (PREC), recall (REC), and the F1 score (F1), are used as part of the analysis. Since the collected dataset does not contain a severe imbalance, accuracy indicates overall model performance as it considers the number of correctly predicted samples out of the entire population. The precision metric indicates the ratio of correctly predicted defective samples to those predicted as defective. This metric is beneficial since detecting defective samples is a priority. The recall metric defines the probability of detection as the ratio of correctly predicted defective samples compared to all truly defective samples. Finally, the F1 metric was selected, being the harmonic mean of the recall and precision metrics.

\subsubsection{Network Resource Efficiency}

The main measure of network resource efficiency used in this work is the average feature throughput, defined as the number of features communicated per second. This metric is proportional to the general throughput metric, which measures the number of bits per second. This work aims to explore the trade-off between the feature throughput and the ML model metrics. The optimal configuration minimizes the feature throughput and maximizes the ML model performance metrics.

\section{Results and Analysis}
This section presents the results obtained from the various experiments conducted. Given the multi-variate nature of the results, stacked parallel coordinate plots, each sharing a common x-axis, are used to display the results of each experiment. The x-axis denotes the list of quantities measured. The y-axis displays the value of each of the measured quantities. The colour of each plotted line represents the ML algorithm used during the experiment. For visualization purposes, the value of the PCA components (denoted by the label PCA\_C) has been scaled for each trial to the range [0.92, 1.00].

\subsection{No PCA}
\begin{figure}[!htbp]
\centerline{\includegraphics[width=0.98\columnwidth]{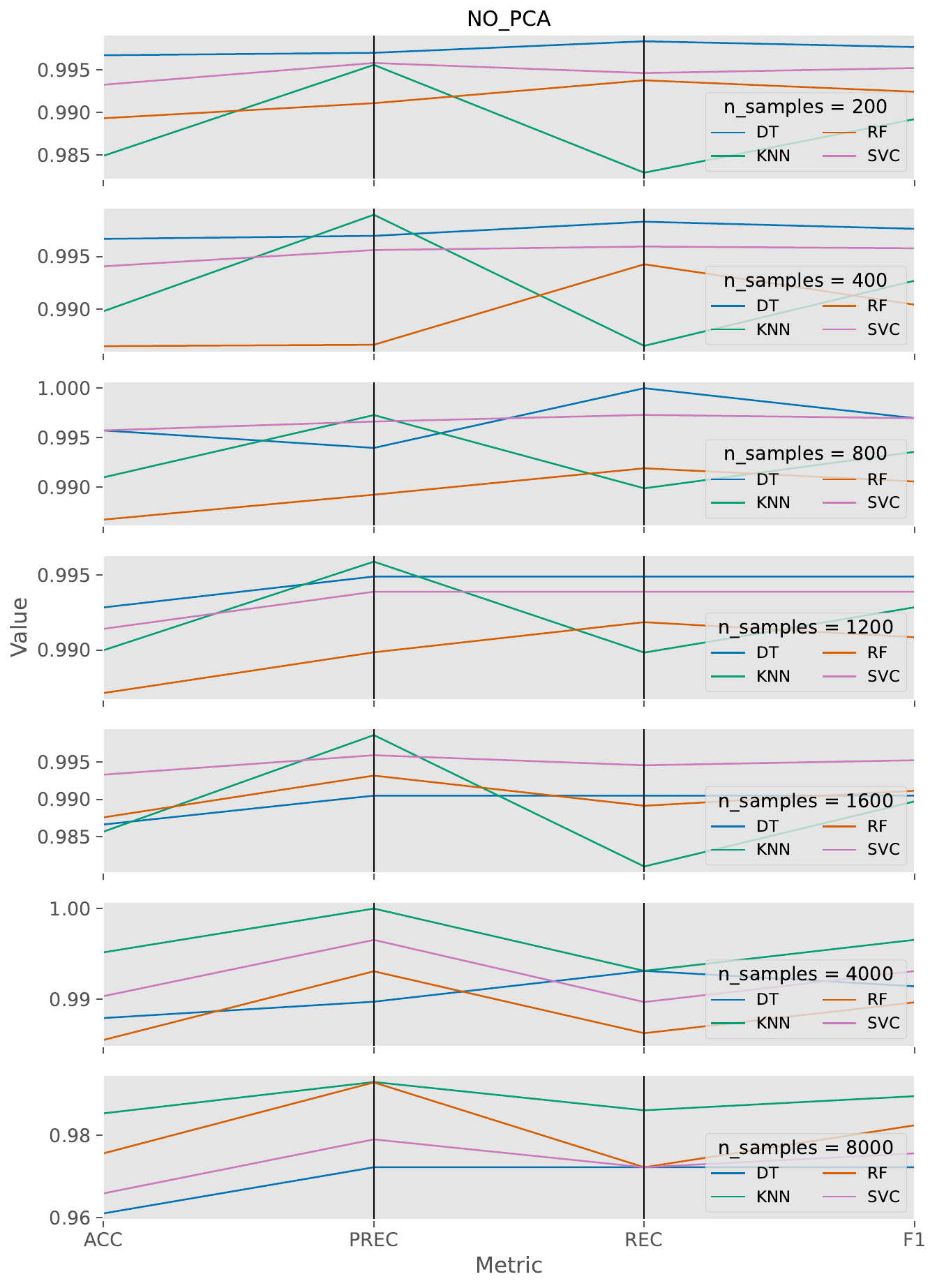}}
\caption{No PCA applied}
\label{no_pca}
\end{figure}
The first set of results, presented in Fig. \ref{no_pca}, correspond to the experiments where no dimensionality reduction was conducted. The $n\_samples$ value in each graph's legends indicates the aggregation interval. As seen, the aggregation interval significantly affects the performance of the ML models. When using a lower aggregation interval, the DT algorithm performs consistently well across all 4 ML metrics and is the best-performing algorithm; however, as the interval is increased, it exhibits increasingly worse performance. Conversely, the KNN algorithm at the lower intervals performed poorly but exhibited increasingly better performance at the higher intervals. In general, the performance of the ML models decreased as the aggregation interval increased, with some algorithms better suited to the smaller and larger intervals, respectively. \par
Regarding communication efficiency, the smallest interval had a feature throughput of 37,280 \textit{features/s}, and the largest interval had a feature throughput of 24,940 \textit{features/s}. Considering both metrics, the best performance for this set of experiments is achieved by the DT and SVC algorithms with an aggregation interval of 800 samples. Unfortunately, neither set of metrics is ideal in this case as excessive amounts of features are relayed, and sub-optimal performance is achieved.

\subsection{STFT PCA}
\begin{figure}[!htbp]
\centerline{\includegraphics[width=0.98\columnwidth]{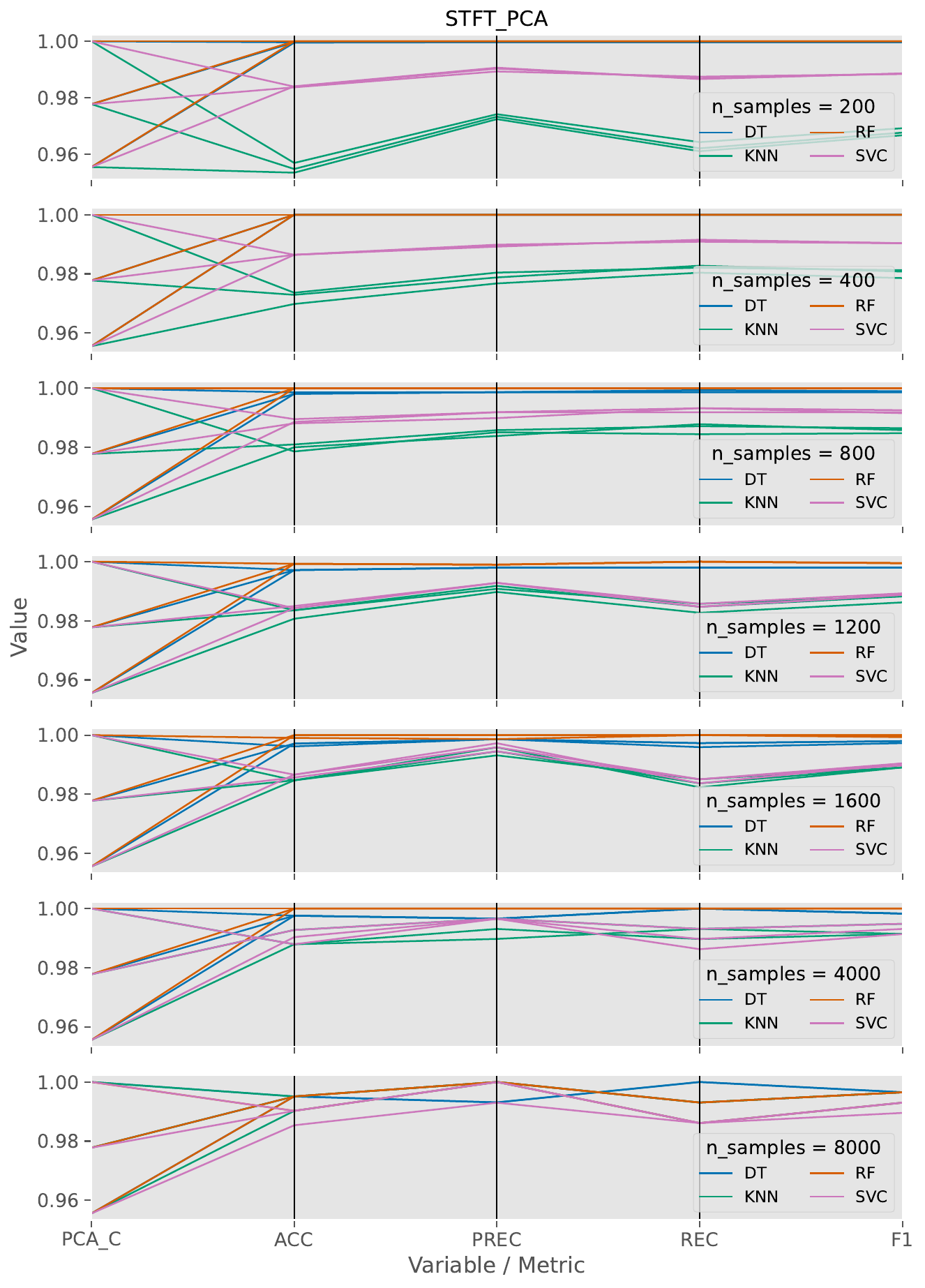}}
\caption{PCA applied to STFT features}
\label{stft_pca}
\end{figure}

In the second set of experiments, PCA was applied to the STFT feature set since it is the most extensive feature set and scales with the aggregation interval. The number of non-STFT features is constant for each feature set at 84. This means that most features are attributed to the STFT feature extraction. The communication efficiency will be significantly improved by using dimensionality reduction on this feature set. Additionally, reducing the feature set will likely positively affect the ML model performance by reducing model complexity, eliminating irrelevant features, and reducing noise. Results from this experiment are present in Fig. \ref{stft_pca}. The number of PCA components used for these experiments was 10, 15, and 20. A noticeable improvement in ML model performance is observed with the tree-based DT and RF algorithms, which consistently exhibit near-optimal performance in most trials. Specifically, the RF algorithm is clearly the best-performing model, with metric values reaching perfect 1.0 scores commonly seen. When the aggregation interval is set at 4000 samples, all PCA trials result in the RF exhibiting perfect performance. Considering this, the RF model's feature throughput when ten principal components are used is 18.8 \textit{features/s}, resulting in a substantial improvement over the previous experiment. Specifically, the use of PCA reduced the best possible feature throughput from the no PCA experiments by 99.92\% while maximizing the ML model performance to the optimal value.

\subsection{All PCA}
\begin{figure}[!htbp]
\centerline{\includegraphics[width=0.97\columnwidth]{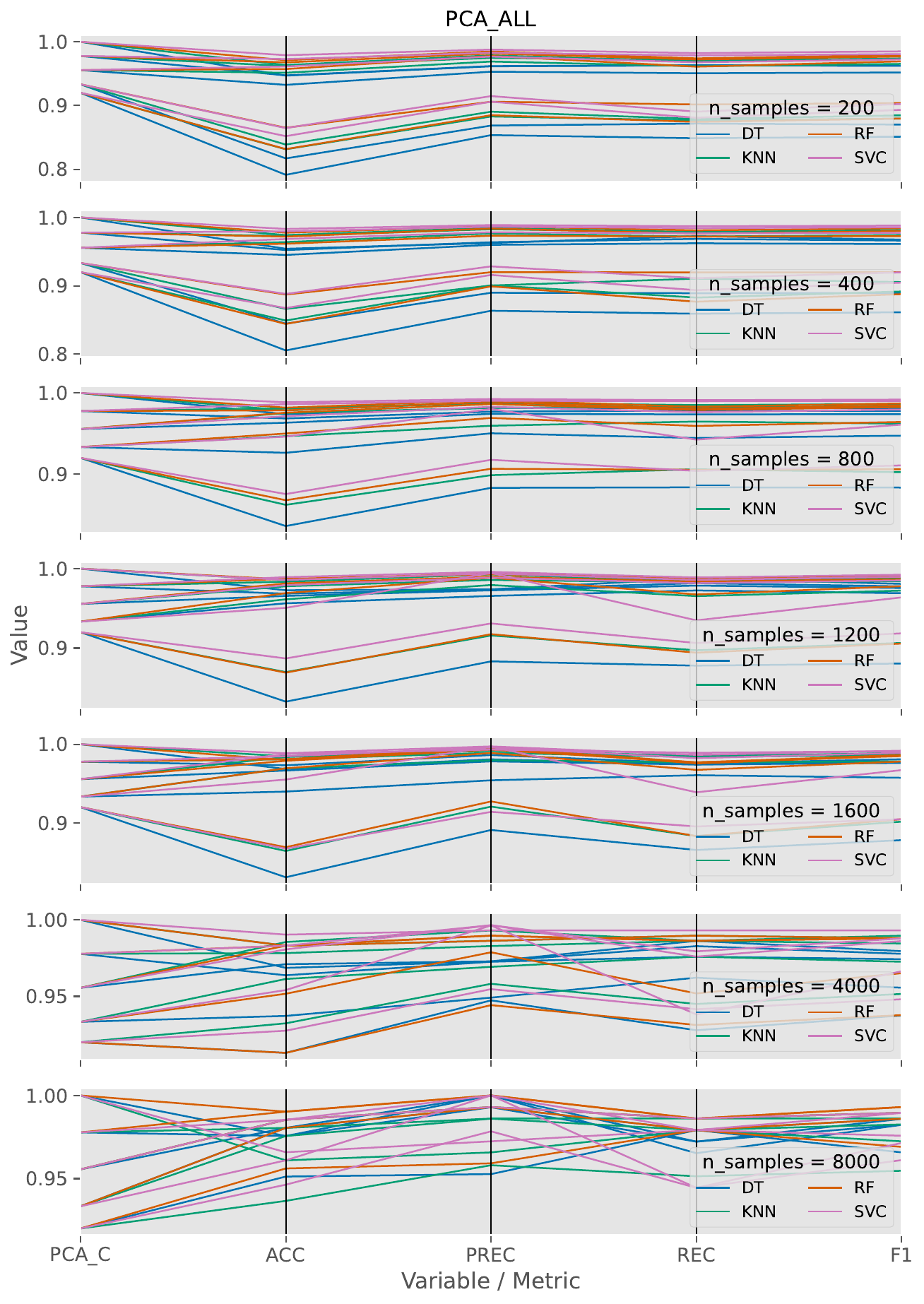}}
\caption{PCA applied to all features}
\label{all_pca}
\end{figure}

The final set of experiments attempt to further improve communication efficiency by applying PCA to the entire feature set, results of which are found in Fig. \ref{all_pca}. The number of PCA components used for these experiments was 2, 5, 10, 15, and 20. The reduction of the entirety of the feature set results in sub-optimal defect detection. Theoretically, the optimal feature throughput achievable in these experiments would be when two principal components were used with an aggregation interval of 8,000 samples, resulting in a feature throughput of 0.2 \textit{features/s}. Based on the results, the ideal trade-off between communication efficiency and model performance occurs when using the RF algorithm, with PCA applied to the STFT feature subset, at an aggregation interval of 4000.

Regarding a comparison with the state-of-the-art, few works consider condition monitoring of a UAV. Of these works, most do not consider network resource optimization. In this regard, a relevant study is conducted by Bondyra \textit{et al.} \cite{bondyra2017fault} which considers 8-element feature vectors and a Support Vector Machine predictor to explore the effect of the data buffer on predictive performance. The feature vectors independently consider FFT, Wavelet Packet Decomposition, or Band Power features. Their method achieved a maximum correct detection ratio of 90\%. The presented results outperform this baseline in terms of accuracy as various feature subsets are considered when making the prediction. Additionally, dimensionality reduction techniques ensure that the input is comparable in scale to the 8-feature element vector to mitigate the trade-off between input feature space and predictive performance.




\section{Conclusion and Future Work}

As demonstrated in this work, developing a practical UAV CM framework has many aspects to consider. One of the critical considerations is the type of model and the performance it can achieve. Another critical consideration is the amount of valuable networking resources consumed as information is relayed through the system. This work determined the optimal configuration parameters that jointly optimize the network resource efficiency and model performance. Results show that using dimensionality reduction techniques reduces the resources consumed while improving model performance. \par
The results presented in this paper pertain to the specific drone, blade type, sensor type/configuration, and environment considered. The drone in this work is a quadcopter. In past work, a feature importance analysis was conducted to determine each feature's contribution to the predicted output. Other quadcopters with similar sensors and sensor deployments are expected to exhibit similar feature importance when contributing to the predictions under the same conditions. The presented work regarding resource optimization also addresses instances where other sensors could be used with different sampling rates. Future work will examine the environment's effect on the solution's robustness. Additionally, a generalization study will be conducted to determine if the observations in this work are transferrable to other drone types and sensor configurations.

\section*{Acknowledgment}
The authors would like to thank Eugen Porter from the Western Engineering Electronics Shop for his help with the drone circuitry and data transmission processes.

\bibliographystyle{IEEEtran}
\bibliography{sample}

\end{document}